%% file: robotac21_llach.tex
\documentclass[letterpaper, 10 pt, conference]{ieeeconf}  

\IEEEoverridecommandlockouts                              

\usepackage{graphicx} 
\usepackage[utf8]{inputenc}
\usepackage[T1]{fontenc}
\usepackage{subcaption}
\usepackage{amsmath} 
\usepackage{amssymb}  

\usepackage{xcolor}
\usepackage[style=ieee,backend=biber]{biblatex}
\bibliography{bibl}

\title{\LARGE \bf
TIAGo RL: Simulated Reinforcement Learning Environments with Tactile Data for Mobile Robots
}

\author{Luca Lach$^{1,2}$,  Francesco Ferro$^{1}$ and Robert Haschke$^{2}$
\thanks{\hrule width0.3\linewidth height 0.8pt}%
\thanks{$^{1}$PAL Robotics S.L., Barcelona, Spain}%
\thanks{{\small\tt luca.lach@pal-robotics.com}}%
\thanks{$^{2}$Neuroinformatics Group, Bielefeld University, Germany}%
\thanks{This work was supported by the European Union Horizon 2020 Marie}
\thanks{Curie Actions under Grant 813713 NeuTouch.}}%

\begin{document}

\maketitle
\thispagestyle{empty}
\pagestyle{empty}

\begin{abstract}
Tactile information is important for robust performance in robotic tasks that involve physical interaction, such as object manipulation.
However, with more data included in the reasoning and control process, modeling behavior becomes increasingly difficult.
Deep Reinforcement Learning (DRL) produced promising results for learning complex behavior in various domains, including tactile-based manipulation in robotics.
In this work, we present our open-source reinforcement learning environments for the TIAGo service robot.
They produce tactile sensor measurements that resemble those of a real sensorised gripper for TIAGo, encouraging research in transfer learning of DRL policies.
Lastly, we show preliminary training results of a learned force control policy and compare it to a classical PI controller.
\end{abstract}

\input{sections/1_introduction.tex}
\input{sections/2_related_work.tex}
\input{sections/3_environments.tex}
\input{sections/4_experiments.tex}
\input{sections/5_conclusion.tex}

\renewcommand*{\UrlFont}{\rmfamily}
\printbibliography

\addtolength{\textheight}{-12cm}   

\end{document}

%% file: sections/1_introduction.tex
\section{INTRODUCTION}

During object manipulation, tactile sensation is one of the most important sensory information for humans.
We utilize it to sense the acquisition of object contacts and to estimate and regulate grasp force.
Additionally, we can detect suboptimal grasp configurations and correct these quickly, for example in the case of object slippage.

Consequently, tactile sensors gained popularity in the research field of robotic manipulation as well.
Many traditional grasping approaches that are devoid of tactile sensations follow a similar basic scheme: objects are segmented on a depth image, grasps are calculated, a trajectory towards the best grasp is computed and finally executed.
These static, open-loop grasping frameworks work well in very controlled environments, however, with increasing uncertainty, their performance can decline sharply.
This is especially the case for mobile service robots as their target environments are typically households, which tend to be unstructured and highly uncertain.

With the addition of tactile sensors to mobile robots, modeling grasping behavior becomes increasingly difficult.
Whereas open-loop grasping systems often consist of static expert systems, approaches using tactile sensors need to model a more dynamic and reactive behavior.
In recent years, Deep Reinforcement Learning (DRL) has been shown to be a powerful tool for learning complex behaviors.
It achieved successes in many domains, for example in video games \cite{lee2018modular}, chess \cite{silver2018general} and also tactile-based robotic tasks \cite{church2020deep}.

As the research field of mobile robotics with tactile sensors in DRL is still growing, basic tools and benchmarks are yet to be defined.
Our contribution in this work is the open-source release of tactile-enabled mobile robotics environments for reinforcement learning.
Specifically, we model PAL Robotic's TIAGo platform \cite{pages2016tiago} with a sensorised end-effector that is also available on the real robot.
The simulated tactile data is transformed to optimally resemble real sensor data in order to facilitate exploration in simulation and sim-to-real transfer learning.
First, we will describe the specification of different environments, explain how we compute the simulated tactile data, and finally show preliminary training results of a force control policy.

%% file: sections/2_related_work.tex
\section{RELATED WORK}
\label{sec:relWork}

In \cite{church2020deep}, a robot arm equipped with a custom-made tactile fingertip learns to type on a Braille keyboard using DRL.
The authors also note the lack of benchmark environments in the domain and propose their task and environment as one for image-based tactile sensors.
Their policy was trained in simulation for some tasks and on the physical robot in others, and they emphasize that transfer learning is an interesting research direction.

Another interesting work that investigates sim-to-real transfer in tactile-based DRL is \cite{wu2019mat}.
By choosing input features that exhibit small differences between simulation and real world, they were able to transfer a policy from simulation to the real robot without relearning.
Their approach involved the conversion of forces to binary forces.
Thus, the tactile data would only indicate object contact, thereby avoiding the demanding task of accurately modeling tactile data in simulation.
Their final model drastically outperformed visual open-loop grasping pipelines with increasing calibration noise.

The authors of \cite{melnikTactile} investigated the benefit of using tactile features when learning dexterous manipulations tasks in simulation.
Their results indicated faster convergence and higher sample efficiency when using tactile data.
These results were similar when using binary and continuous tactile data.
The environments they developed were published and made open-source as part of OpenAI's gym \cite{gym}.

\cite{huang2019learning} used a Shadow Hand equipped with BioTac sensors to learn gentle object manipulation.
They conducted experiments in simulation as well as on a real robot, however, they do not attempt transferring policies.
Nonetheless, their work is interesting to mention as their objective was to learn gentle force control.
They achieve this by shaping a force-based reward function and comparing different alternatives.

%% file: sections/3_environments.tex
\section{REINFORCEMENT LEARNING ENVIRONMENTS}

In reinforcement learning, an agent attempts to learn a policy that maximizes the expected reward in an environment.
Therefore, it receives observations and in turn, needs to compute an action to take.
The environment itself models its dynamics internally, hidden from the agent, only providing observations.

TIAGo RL\footnote{\url{https://github.com/llach/tiago_rl}} comprises three different base environments, two with tactile sensors and one without.
Similar to OpenAI gym's \cite{gym} robotics environments, the idea was to model all common aspects of an environment in its base class, e.g.\ the robot model or joint limits.
Task-specific environments that inherit from such a base class can simply override environment properties such as the reward function or the task setup and thus avoid re-implementing basic robot dynamics.
Additionally, all derived task environments benefit from bug fixes or improvements in the base environment.
We use PyBullet \cite{pybullet}, a free and open-source real-time physics simulator, to simulate the environment's dynamics.
PyBullet also allows the stiffness and damping parameters of collision bodies to be modified, making it attractive for learning of force control policies.

By default, all environments model a grasping scenario where the robot is located next to a table with an object.
The robot's arm and gripper are configured in a pre-grasp pose, meaning object contact is anticipated when the gripper is closed.

\begin{figure}
    \centering
    \subfloat[\emph{TIAGoTactileEnv}]{\label{fig:envRobot}
        \includegraphics[width=0.49\linewidth]{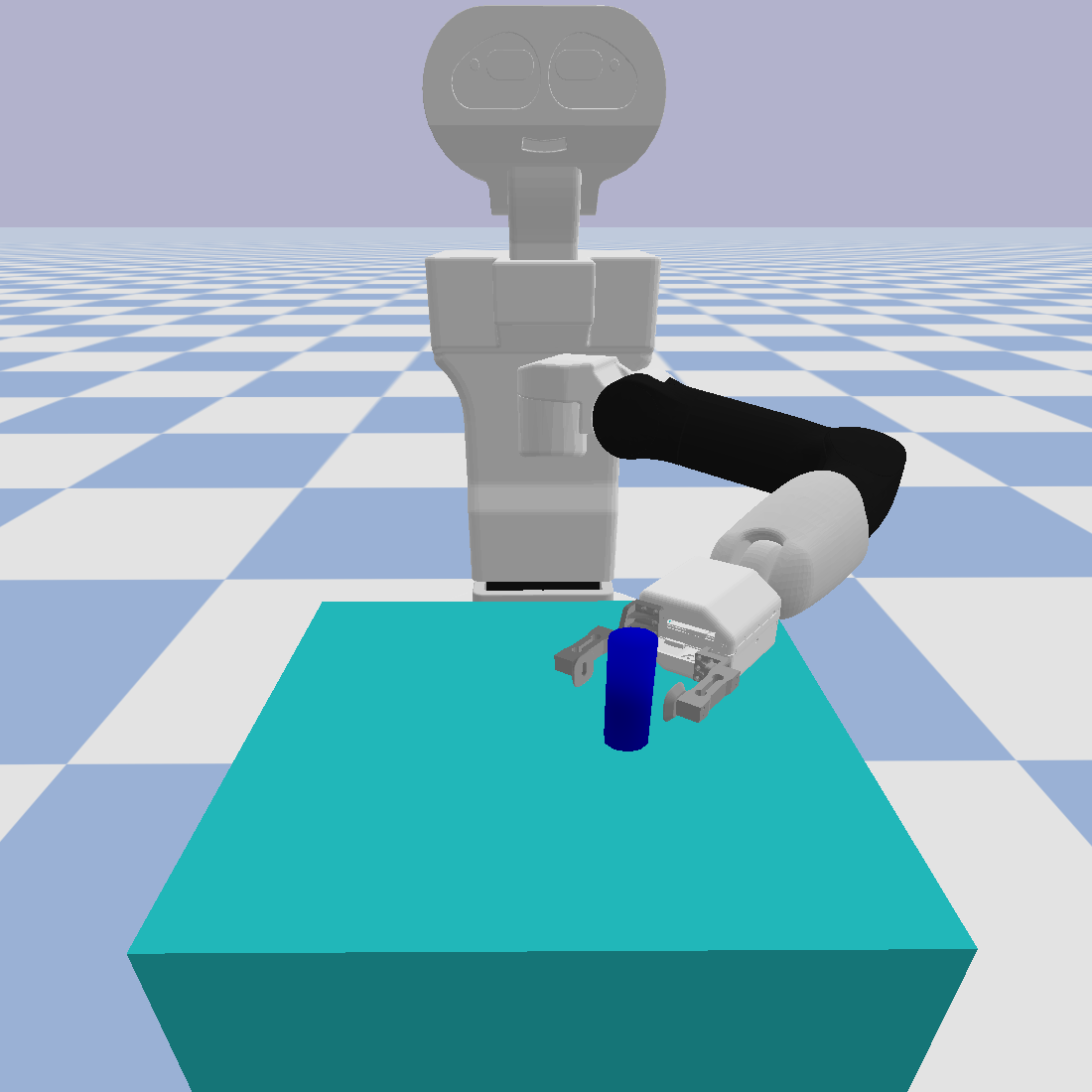}}
    \subfloat[\emph{GripperTactileEnv}]{\label{fig:envGripper}
	   \includegraphics[width=0.49\linewidth]{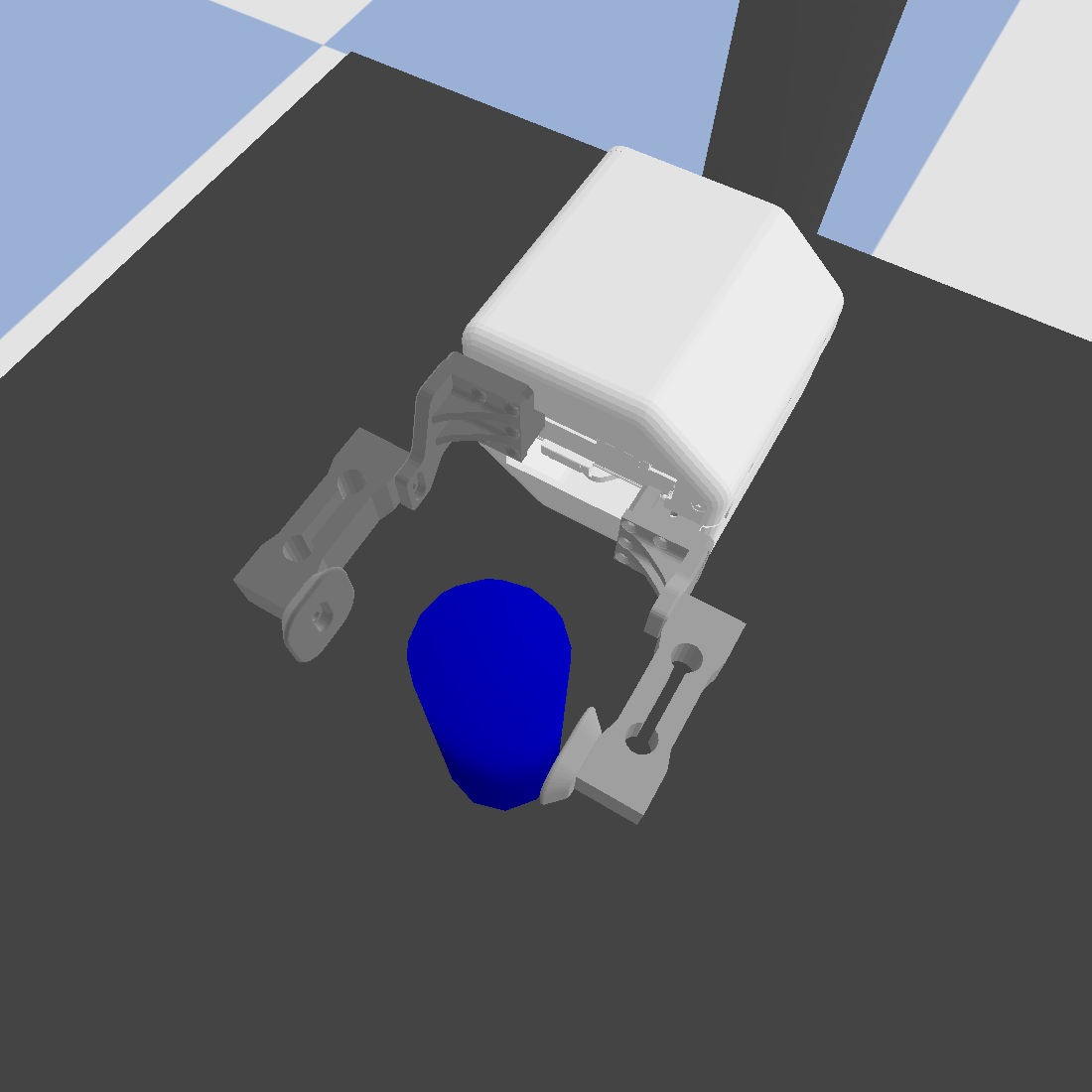}}
    \caption{Initial configurations of both tactile environments with the gripper in a pre-grasp pose. The object is closer to one of the fingers, rather than being centered perfectly.}
    \label{fig:environments}
\end{figure}

\subsubsection{TIAGoTactileEnv}
This environment uses the default TIAGo robot model with the exception that the gripper's fingers are replaced by TA11 load cell sensors.
It replicates a benchmarking setup for tactile-based grasping that can be evaluated on TIAGo.
Results obtained in this environment can be easily compared to those of one without tactile sensors to evaluate their benefits.

\subsubsection{GripperTactileEnv}
In this environment, only the gripper model with load cell sensors is used instead of the complete robot.
The gripper is fixed above the table in a pre-grasp position as described above.
It was created to learn gripper-only control policies for scenarios where the reaching motion of the arm is computed by a classical kinematics solver.
The part of the manipulation task where physical contact is expected can then be handled by the gripper-only policy.
By removing many collision bodies from the environment, we expect this environment to be less computationally demanding.
Moreover, the DRL algorithm does not need to learn to ignore the arm's degrees of freedom and should thus converge faster.

\subsubsection{TIAGoPALGripperEnv}
Lastly, we also include a sensorless environment as there is currently no open-source DRL environment available for TIAGo.
Hence, it can be used as a benchmark for \emph{TIAGoTactileEnv} and to replicate tasks from gym's robotics environments.
Note, that using this environment is different from using \emph{TIAGoTactileEnv} and ignoring the tactile data because the robot models differ.

\subsection{Tactile Data Modeling}
\label{sec:tacDataModel}

\begin{figure}
    \centering
    \subfloat[Force values while grasping with TA11 sensors]{\label{fig:realForce}
	   \includegraphics[width=0.9\linewidth]{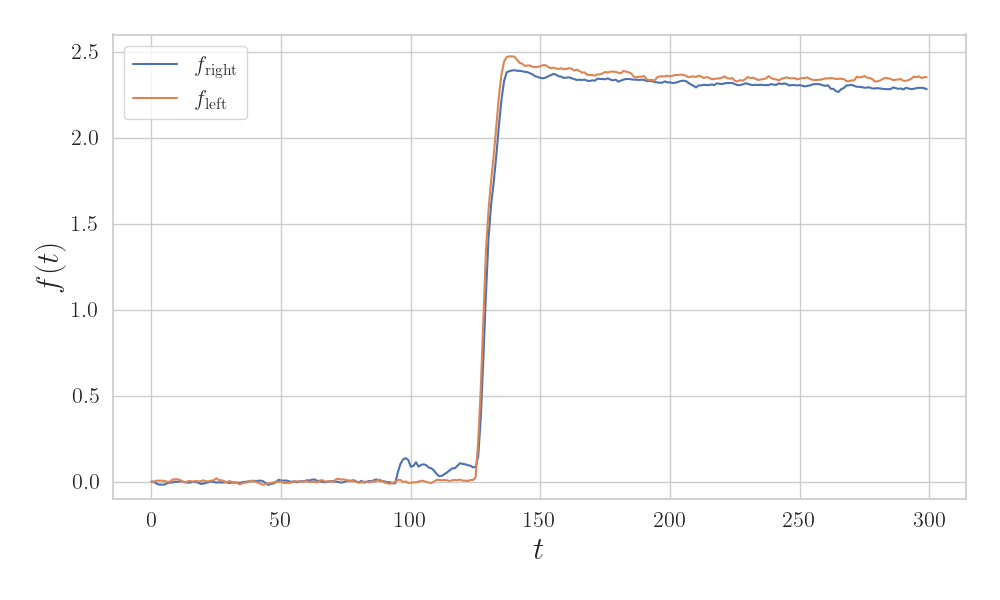}}\\
    \subfloat[Simulated force values $f^{\text{raw}}$, calculated using equation \ref{eq:forceTrans}]{\label{fig:simForce}
	   \includegraphics[width=0.9\linewidth]{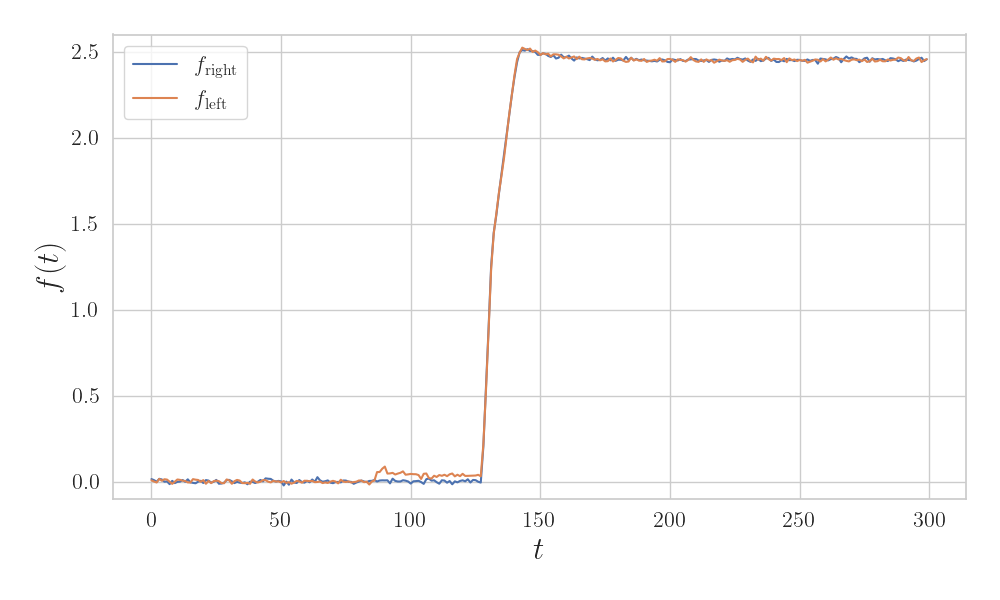}}
    \caption[Force value comparison]{Comparison of force values measured while grasping. \ref{fig:realForce} shows the force trajectory of grasp performed on a real TIAGo. \ref{fig:simForce} shows the simulated forces of the same grasp replicated with our environment.}\label{fig:forceComp}
\end{figure}

One important part of the two \emph{TactileEnvs} is how they model tactile data.
We recorded the force values of several object grasps on a real robot and in simulation.
The object's weight, its position relative to the gripper, and finger velocity were identical to the real-world values.
In PyBullet, we extracted the contact forces $f^{\text{contact}}$ between each finger and the object.
We found that scaling $f^{\text{contact}}$ with a factor of 100 would result in force ranges that are very similar to the ones measured on the real robot.
Furthermore, we estimated the real sensor's standard deviation in non-contact state using $10.000$ samples, yielding $\sigma = 0.0077$.
The simulated force output for sensor $i$ is eventually computed as follows:
\begin{equation}
    \label{eq:forceTrans}
    \begin{gathered}
        f^{\text{raw}}_i(t) = f^{\text{contact}}_i(t) * 100 + \epsilon \\
        \text{with}\,\, \epsilon \sim \mathcal{N}(0, 0.0077)
    \end{gathered}
\end{equation}
Note, that we refer to these forces as "raw" as they simulate raw sensor readings on the robot despite the transformations we apply in simulation.
Figure \ref{fig:forceComp} compares two force trajectories of grasps performed in simulation and on a real robot.

Binary forces were found to facilitate transfer learning as reported by \cite{wu2019mat}.
Hence, our environments offer this conversion of force values as well.
In this representation, the tactile data solely indicates object contact without information about its intensity.
It is calculated with:
\begin{equation}
    \begin{gathered}
        f^{\text{binray}}_i(t) =
        \begin{cases}
            1,      & \text{if } f^{\text{raw}}_i(t) > f^{\text{thresh}}_i\\
            0,      & \text{otherwise}
        \end{cases}
    \end{gathered}
\end{equation}
where $f^{\text{thresh}}_i$ is a sensor-specific noise threshold.

\subsection{Observation Space}

At each timestep, the agent makes an observation $o(t)$ which is then used to calculate a subsequent action.
The base environment without tactile sensors, \emph{TIAGoPALGripperEnv}, provides the agent with information about the joint states:
\begin{equation}
    o^{\text{base}}(t) = \left(
        \begin{array}{c}
               q_1(t)\\
               \vdots\\
               q_{N}(t)\\
               \dot{q}_1(t)\\
               \vdots\\
               \dot{q}_{N}(t)\\
        \end{array}
  \right)
\end{equation}
where $q_j(t)$ and $\dot{q}_j(t)$ refer to the position and velocity of joint $j$ at time $t$ respectively.
For TIAGo we have $N=10$ joints, including seven arm joints, two gripper joints, and the torso lift joint.
Any task-specific environment would need to append some information relevant to the task's goal to these basic observations, e.g.\ distance to a certain goal position.

For the two tactile environments, the current force deltas are also included in the observations:
\begin{equation}
    o^{\text{tactile}}(t) = \left(o^{\text{base}}(t), \Delta f_{\text{right}}, \Delta f_{\text{left}} \right)
\end{equation}
with $\Delta f_{\text{i}} = f_{\text{i}}(t)-f^{\text{goal}}$.
While for \emph{TIAGoTactileEnv}, $N=10$, the gripper-only environment \emph{GripperTactileEnv} only actuates the two gripper joints, hence $N=2$.

\subsection{Action Space}
The action simply consists of the desired velocities for each joint:
\begin{equation}
    a(t) = \left(
        \begin{array}{c}
               \dot{q}^{\text{des}}_j(t)\\
               \vdots\\
               \dot{q}^{\text{des}}_{N}(t)\\
        \end{array}
  \right)
\end{equation}
The action space is designed to mimic the expected input for TIAGo's position controller, another property aimed at facilitating easy policy transfer.
Moreover, joint velocities are restricted through PyBullet to not surpass the same limits that are in place on the real robot as a measure to prevent the policy from learning unrealistic behavior.

\subsection{Reward Function}

The reward at time $t$ is given by:
\begin{equation}
    r(t) = - \!\!\!\!\!\! \sum_{i \in \{\text{right}, \text{left}\}} \!\!\!\!\!\! | f_i(t) - f^{\text{goal}} |
\end{equation}
where $f^{\text{goal}}$ is a goal force that should be reached and maintained.
The intention behind modeling the reward function like this is to encourage the agent to maintain a particular goal force.
By taking the absolute value, the agent is equally penalized for forces above and below the goal force.
Ultimately, the agent needs to learn a force control policy in order to maximize its reward.

%% file: sections/4_experiments.tex
\section{LEARNING FORCE CONTROL}

We conducted a preliminary experiment to assess the feasibility of learning force control policies using our environments.
As our goal is to demonstrate policy learning in situations where tactile data is most relevant, we chose to use the \emph{GripperTactileEnv} for the experiment.
In the learning scenario, the gripper starts in an open position.
An object is placed centered between the fingers, thus guaranteeing object contact when the fingers are closed.
The \emph{stiffness} and \emph{damping} parameters were set to simulate a slightly compliant object.
The object's weight and size were set to match those of a real object we used in a similar real-world grasping experiment.

We trained the agent using the Twin Delayed Deep Deterministic Policy Gradient
(TD3) \cite{td3} algorithm.
It can learn continuous actions, is an off-policy algorithm, and performed very well in different kinds of environments, including robotics.
As in the original paper, we used Gaussian noise $\mathcal{N}(0, 0.1)$ for policy exploration.
All other training parameters were left unchanged.

\begin{figure}[h!]
    \centering
    \includegraphics[width=\linewidth]{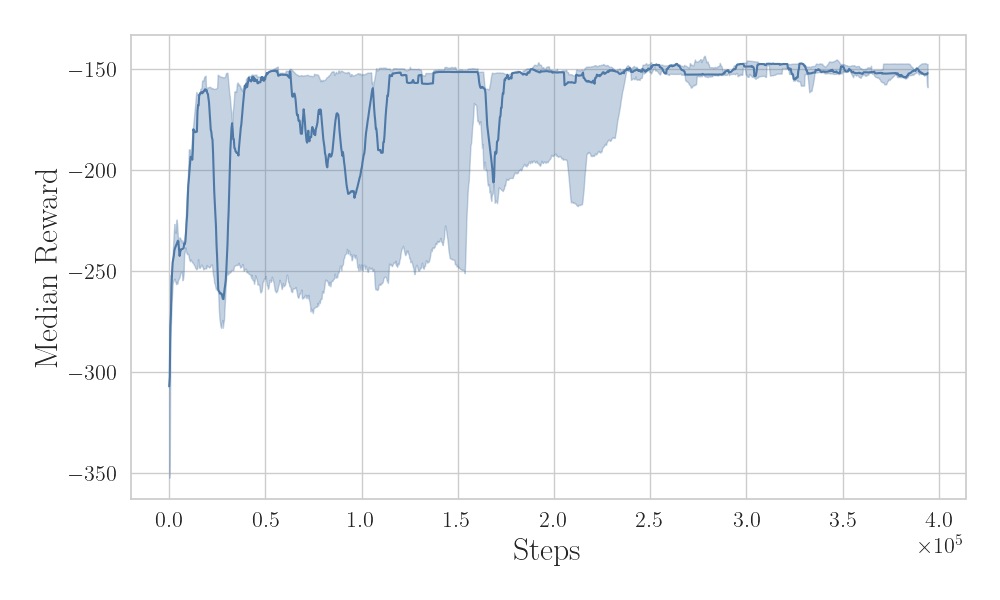}
    \caption{Median reward of policies trained with TD3 on \emph{GripperTactileEnv} across five different seeds.}
    \label{fig:train}
\end{figure}

The agent was trained for $4 \times 10^5$ timesteps with an episode length of $300$ steps.
Once the end of an episode was reached, the environment resets itself and the agent could try anew.
During training, the mean reward over the last 20 episodes was periodically recorded.
On occasions where the policy had surpassed its former best mean reward, the policy's weights were saved as well.
Figure \ref{fig:train} shows the median reward over five different random seeds.
During the first part of the training, the agent steadily achieved higher rewards until it converged to a maximum of about $-150$.
After around $2.5 \times 10^6$ steps, the performance variance across seeds stayed relatively low, indicating that all agents converged to similar solutions.

\subsection{Force Control Policy Evaluation}
In order to evaluate the agent, we compared it to a classical force controller that we usually employ on TIAGo.
This controller starts by closing the fingers using standard position control.
If a finger detects a force above the noise threshold, it stops moving until the opposing finger acquires contact as well.
Thereby it avoids undesired object movements.
Once both fingers touch the object, the controller switches to force control and tries to minimize the difference of measured force to goal force.
It uses PI control to reach and maintain the goal force.
The integral part of the controller compensates for inaccurate estimations of the object's stiffness.

For the evaluation, we repeated the experiment 10 times with this classical force controller and a trained agent.
We report the mean reward and its standard deviation from the trials for each method.
The force controller achieved a mean episode reward of $-198.01 \pm 14.52$ while the agent achieved $-146.93 \pm 2.68$.
From a reward perspective, the agent outperformed the PI force controller. However, from observing the agent's behavior, it became clear that it could not be transferred to a real robot:
It learned to bounce the object back and forth between the fingers, thus creating short pulses of impact force reducing the reward rather than penetrating the object with a stable grasp and maintaining grasp forces.

%% file: sections/5_conclusion.tex
\section{CONCLUSION}

In this work, we have introduced novel reinforcement learning environments for TIAGo that incorporate tactile data.
Two of the presented environments provide tactile data from simulated load cell sensors, the third one without tactile sensors can be used to benchmark the advantage of using tactile data.
The simulated sensor data is transformed to match the characteristics of real sensor data by scaling it and applying Gaussian noise.
In preliminary experiments, an agent has shown the ability to learn to reach and maintain a certain goal force in our environments.
However, it seemed to have overfitted to that particular training scenario.

The source code is being actively developed, maintained and new features are continuously added.
In future works, we plan to investigate how to learn more reliable force control agents in simulation that generalize better across different goal forces.
Possible approaches include refining the reward function and hyperparameter searches.
Ultimately, the goal is to research transfer learning capabilities using these simulation environments.